\documentclass{article}

     \PassOptionsToPackage{numbers, compress}{natbib}

 \usepackage{amsmath}

 \usepackage[dblblindworkshop, final]{neurips_2025}
\workshoptitle{Imageomics Workshop}

\usepackage[utf8]{inputenc} 
\usepackage[T1]{fontenc}    
\usepackage{hyperref}       
\usepackage{url}            
\usepackage{booktabs}       
\usepackage{amsfonts}       
\usepackage{nicefrac}       
\usepackage{microtype}      
\usepackage{xcolor}         
\usepackage{graphicx}
\usepackage{float}

\title{ProteinPNet: Prototypical Part Networks for Concept Learning in Spatial Proteomics}

\author{%
  Louis McConnell \\
  Lausanne University Hospital \\
  Lausanne, CH \\
  \texttt{louie.mc@berkeley.edu} \\
  \And
  Jieran Sun\\
  Lausanne University Hospital \\
  Lausanne, CH \\
  \texttt{jieran.sun@chuv.ch} \\
  \And
  Theo Maffei \\
  Lausanne University Hospital \\
  Lausanne, CH \\
  \texttt{theo.maffei@chuv.ch} \\
    \And
  Raphael Gottardo \\
  Lausanne University Hospital \\
  Lausanne, CH \\
  \texttt{raphael.gottardo@chuv.ch} \\ 
  \And
  Marianna Rapsomaniki \\
  Lausanne University Hospital \\
  Lausanne, CH \\
  \texttt{marianna.rapsomaniki@chuv.ch} \\ 
}

\begin{document}

\maketitle

\begin{abstract}
Understanding the spatial architecture of the tumor microenvironment (TME) is critical to advance precision oncology. We present \textbf{ProteinPNet}, a novel framework based on prototypical part networks that discovers TME motifs from spatial proteomics data. Unlike traditional post-hoc explanability models, ProteinPNet directly learns discriminative, interpretable, faithful spatial prototypes through supervised training. We validate our approach on synthetic datasets with ground truth motifs, and further test it on a real-world lung cancer spatial proteomics dataset. ProteinPNet consistently identifies biologically meaningful prototypes aligned with different tumor subtypes. Through graphical and morphological analyses, we show that these prototypes capture interpretable features pointing to differences in immune infiltration and tissue modularity. Our results highlight the potential of prototype-based learning to reveal interpretable spatial biomarkers within the TME, with implications for mechanistic discovery in spatial omics\footnote{Code available at \url{https://github.com/AI4SCR/ProteinPNet}.}.
\end{abstract}

\section{Introduction}

Tumors are complex ecosystems where diverse cell populations interact to form heterogeneous tumor microenvironments (TMEs) \cite{kashyap_quantification_2022}. The spatial heterogeneity of the TME has been the focus of intensive research, enabled by a range of tumor profiling technologies. From hematoxylin and eosin (H\&E) staining, routinely used to assess morphological tissue alterations, to single-cell technologies that capture full molecular profiles, emerging data are starting to uncover prognostic patterns within the TME \cite{balkwill_tumor_2012}, associated with different spatial patterns of immune cells \cite{abduljabbar_geospatial_2020, gong_quantitative_2019} or cancer-associated fibroblasts (CAFs) \cite{mhaidly_fibroblast_2020, ozdemir_depletion_2014}. Spatial heterogeneity is also associated with differences in tumor architecture, as different morphological structures can lead to diverse cancer invasion patterns \cite{noble_spatial_2021}. Recent developments in single cell and spatial omics technologies now enable the deep molecular profiling of each individual cell within the TME \cite{lewis_spatial_2021}, offering an unprecedented opportunity to address an outstanding question in cancer biology: \textit{how can we discover recurrent spatial patterns within the TME that drive disease outcomes?} Identifying and targeting these spatial biomarkers could spearhead the development of more precise and personalized therapies.

Despite growing data availability, extracting biologically meaningful spatial patterns from spatial omics data remains a challenge. In histopathology, where traditional deep learning models trained on H\&E images have demonstrated high predictive power on a number of clinical tasks \cite{echle_deep_2021, bahadir_artificial_2024}, attention-based multiple-instance learning (MIL) pipelines \cite{ilse2018attention} and post-hoc explainers (e.g., Grad-CAM \cite{selvaraju2017grad}, layer-wise relevance propagation \cite{montavon2019layer}, saliency maps \cite{guo2022attention}), allow visualization of TME regions that influence the models' decisions. In the nascent field of spatial omics, current work on identifying spatial biomarkers is heavily based on well-known tumor properties \cite{jiang_meti_2024}, for example, hand-crafted spatial features of immune cells to predict immunotherapy response \cite{williams_current_2024, hoch_multiplexed_2022, wang_spatial_2023}. Early attempts to devise methods that automatically discover spatial patterns are emerging \cite{wu_graph_2022, tanevski_explainable_2022, zuo_dissecting_2024}. While useful, both histopathology and spatial omics methods for mechanistic discovery are based on diffuse, hard-to-interpret post-hoc explanations that do not necessarily reflect the true mechanistic basis of the models' predictions and generally have problems with faithfulness \cite{rudin2022interpretable, laugel2019dangers}.

An attractive alternative to post-hoc explainers is to build inherently interpretable models \cite{rudin2019stopexplainingblackbox}. Prototypical part networks \cite{chen_this_2019} are one type of interpretable model that learn prototype representations, where each prototype corresponds to a representative part training example, and predictions are made by comparing input images to prototypes. Because the prediction is a function of the computed similarity between prototypes and the input image, they are able to circumvent many of the faithfulness problems of post-hoc explanations. Although prototypical part networks have been used in the medical domain for interpretable, case-based deep learning in clinical applications \cite{noauthor_case-based_nodate, wei_mprotonet_2023, kim_xprotonet_2021}, their potential for mechanistic discovery in the realm of spatial omics is underexplored. These networks are particularly fitting to the problem of identifying spatial biomarkers from spatial omics, as they resemble how oncologists or pathologists reason and use resemblance to learned spatial motifs in the data as a bottleneck layer in prediction. In this paper, we propose \textbf{ProteinPNet}, a novel prototypical part-based model tailored to spatial proteomics data. We show that ProteinPNet learns prototypes that align with distinct tumor subtypes and reflect biologically relevant patterns, as assessed via graphical and morphological analysis. Although preliminary, our findings suggest that prototype-based learning holds considerable promise in identifying spatial biomarkers in the TME and guiding future discoveries in oncology.

\section{Methods}

\begin{figure}[h]
\includegraphics[width=\textwidth]{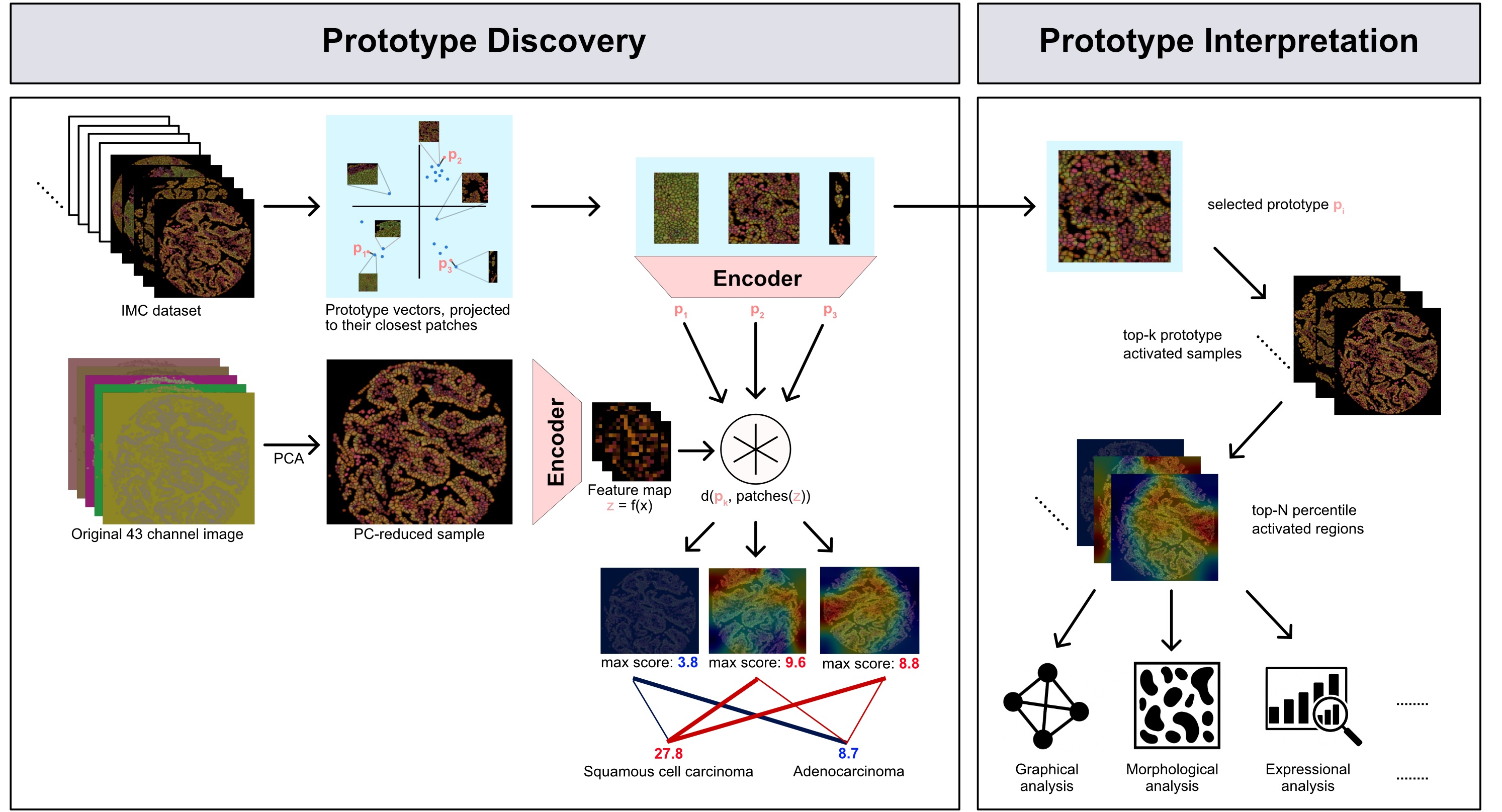}
\caption{\textbf{The ProteinPNet workflow.} During \textit{prototype discovery}, the prototype vectors are randomly initialized and projected onto the closest patch representation. The prototype representations are then convolved over the representation of the spatial proteomics image with a cosine similarity kernel to generate an activation heatmap, which generates a set of prototype scores that are linearly combined to make the final prediction. During \textit{prototype interpretation}, prototypes that generated the highest accuracy are analyzed in terms of their morphological and compositional characteristics.} 
\label{fig1}
\end{figure}

We developed \textbf{ProteinPNet}, a framework for prototypical part learning tailored to spatial proteomics data that enables the discovery of interpretable, morphologically-aware spatial biomarkers. The ProteinPNet framework consists of two principal stages, namely \textit{prototype discovery} and \textit{prototype interpretation} (Figure \ref{fig1}). During \textit{prototype discovery},  a prototypical part network inspired by \citet{chen_this_2019}\footnote{We follow the notation from this paper as well. For more details, please see \cite{chen_this_2019}.} is trained to directly learn prototypical concepts, hereafter referred to as prototypes, from spatial proteomics data. 
To account for the high dimensionality and variable channel structure of spatial proteomics compared to standard RGB images, we treat each protein measurement as an image channel and design a custom encoder architecture. While several options for analyzing data with a large number of input channels are available, we encountered a high amount of overfitting when taking in all protein channels (over 97\% train accuracy and roughly random train performance), even when using small networks (Resnet18 \cite{he_deep_2015}) and low numbers of prototypes (1 per class). Instead, we reduced the data to three principal components (PCs) and used a pretrained ResNet152 backbone. 
%
Similar to \citet{chen_this_2019}, ProteinPNet performs prediction by regressing on a set of instance-specific scores that measure the degree to which a prototype is a part of the instance. For a given spatial proteomics sample $x$ and a convolutional head $\mathbf{z} = f(\mathbf{x})$ with convolutional output representations of dimension $H \times W \times D$, ProteinPNet learns $m$ prototypes $\mathbf{P}=\{{\mathbf{p}_j\}}^{m}_{j=1}$ of shape $H_p \times W_p \times D$, where $H_p \leq H, W_p \leq W$. From the prototypes and a sample z, we compute the prototype scores $g_{p_j}(z) :=  \min_{\tilde{z} \in \text{patches}(z)} d(\tilde{\mathbf{z}}, p_j)$ for some distance metric $d$.  Following \cite{willard2024looksbetterthatbetter}, we use cosine distance as our distance metric to get our prototype scores $g_{p_j}(z)$. The scores are linearly combined to map to the output distribution used for prediction.
To maintain interpretability, at every $k$ epochs, each prototype $p_j$ is assigned to the filter representation of the closest patch $z$. This ensures that each prototype has a concrete visual reference directly aligned with the prototype vector used to compute its scores 
(see \cite{chen_this_2019} for details). 

During \textit{prototype interpretation}, these prototypical concepts are analyzed in terms of their morphological, topological, and protein-expression-level characteristics to reveal differences in the underlying biology and link to different TME topologies, pathways, or niches.
Once learned, the prototypes can be further interrogated using custom downstream analyses to assess if they are driven by differences in tumor morphology, cellular composition, or both. Prototype activation maps allow prototype isolation by selecting the top $N$-percentile most activated regions in the top-$k$ most activated samples. Those regions can then be subjected to domain-specific analyses (e.g., using graph-based or morphological scores, or differential expression analysis), revealing key biological processes. This stage of analysis can be customized to the biological hypothesis and the dataset at hand. 

\vspace{-10pt}

\section{Results}

\textbf{Benchmarking on synthetic data} We first evaluated the effectiveness of ProteinPNet in identifying prototypical motifs on synthetic data with ground-truth prototypes. We generated a synthetic dataset consisting of 3-channel images assigned to two different hypothetical classes, containing different types of randomly distributed three-node subgraphs (Figure \ref{supp:synth_data}), and injected a unique subgraph in each class (red circles). This proxy task is important, as it allows us to evaluate ProteinPNet on ground-truth prototypes which do not exist in real-world spatial omics data. It is thus essential that ProteinPNet can provably recover known class-discriminative synthetic prototypes while ignoring the neutral ones. We assessed ProteinPNet's performance both in terms of accuracy, as well as by checking whether the injected class-specific pattern was detected in the extracted prototypes. ProteinPNet consistently achieved a 100\% classification accuracy (Table \ref{table1}). In every run, prototype activation maps recovered the exact area containing the injected prototype for at least one class (see representative examples in Figure \ref{synth_data_activations}). Interestingly, in one seed, the model used white space to indicate the absence of a key prototype, while in all others, it identified class-specific prototypes in both classes (Figure \ref{extracted_prototypes}). This demonstrates an interesting effect in which ProteinPNet is able to learn prototypes identifying the "absence" of key prototypes. 

\begin{figure}
\includegraphics[width=\textwidth]{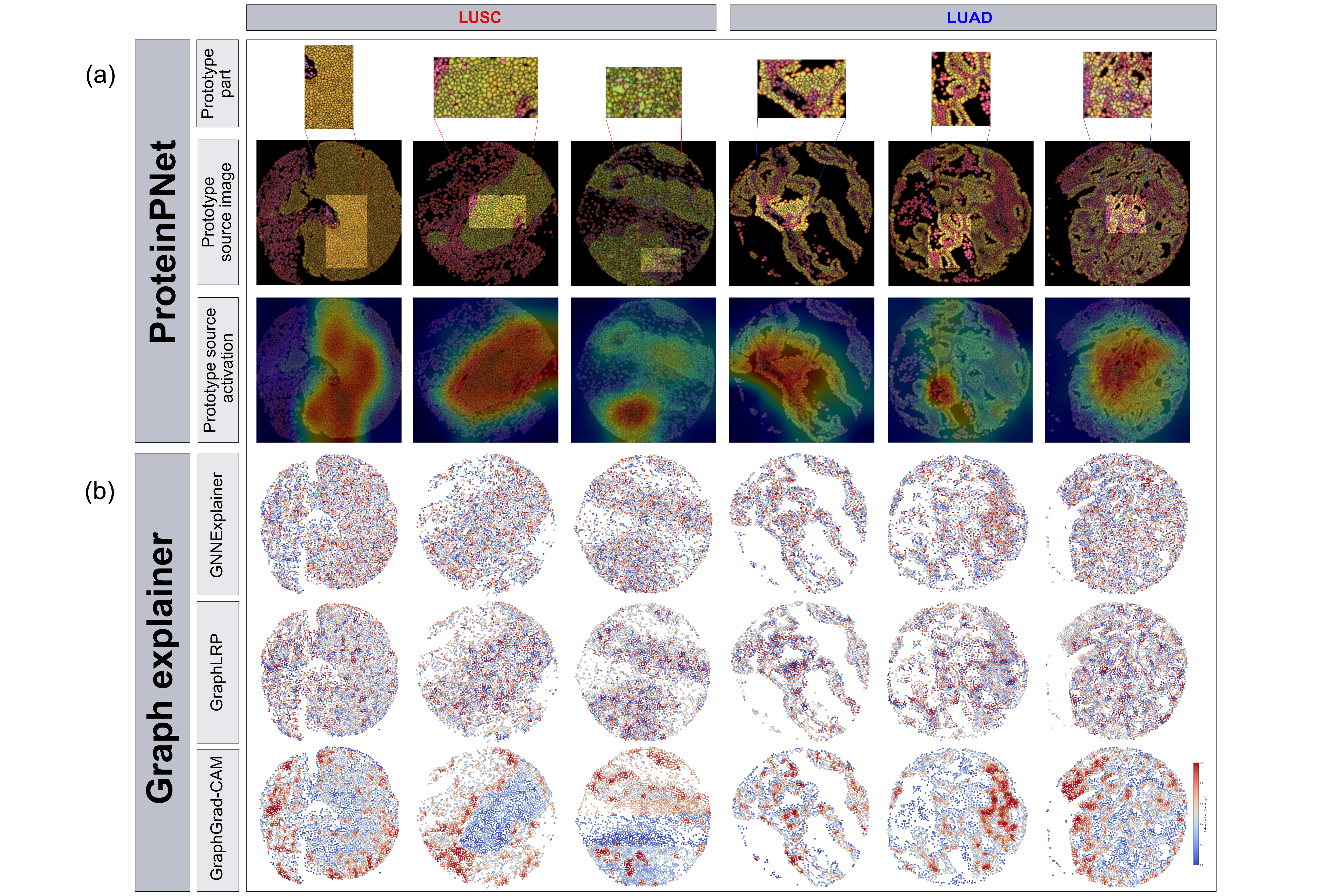}
\caption{(a) Characteristic examples of LUSC and LUAD prototypes, collected across many runs with only one prototype per class, together with the source image and activation maps. (b) Performance of three graph explainers for the same example samples.} \label{learned_prototypes}
\end{figure}

\textbf{Evaluation on real-world spatial proteomics data} 
We then applied ProteinPNet on a publicly available spatial proteomics dataset \cite{cords_cancer-associated_2024}, from a large non-small cell lung cancer (NSCLC) cohort containing both adenocarcinoma (LUAD) and squamous cell carcinoma (LUSC), profiled by imaging mass cytometry. The dataset contains a total of 1021 samples, with each sample corresponding to the simultaneous quantification of 43 different protein markers, resulting in a 43-channel image. 
We first trained ProteinPNet to classify between different NSCLC subtypes, i.e., LUAD vs. LUSC. We note that, across all 1021 samples, 618 belong to the LUAD and 403 to LUSC subtype, giving a naive accuracy of 60.5\% when predicting only the majority class. We used the predictive accuracy of the classifier as an initial validation of the predictive value of the learned prototypes: a low predictive loss in this task suggests that the prototype scores - and consequently the prototypes - contain important information needed to distinguish data between cancer classes. ProteinPNet reached an 80.7\% accuracy on LUAD vs. LUSC prediction (Table \ref{table1}). 
To probe what information the model relies on, we conducted two ablations. First, we evaluated whether learned prototypes outperform random crops; in other words, are the prototype scores indicative of unique prototype selection, or will a distance metric to any random patch have the same predictive power? To test this, we froze the model backbone after training beyond 80\% accuracy and reinitialized the prototype vectors, pushing them onto random patches. This setup ensures that the backbone is capable of representing high quality prototypes while testing the quality of the learned vs. random prototypes. Accuracy dropped by 6.2\% (Table \ref{table1}), indicating that prototypes learned by the model encode meaningful information about the underlying tissue structure. Second, we removed the cell type information by setting all cell representations to a constant, leaving only morphological information. In this setting, we were only able to reach an accuracy of 71.2\%, suggesting that both morphological features and protein expression are essential for prediction.

We then benchmarked ProteinPNet against graph-based explainers. We first applied different types of Graph Neural Networks, namely Graph Convolutional Networks (GCN) \cite{kipf2017semi}, Graph Isomorphism Networks (GIN) \cite{xu2019powerful} and Graph Attention Networks (GAT)\cite{velickovic2018graph}, for the same prediction task (LUAD vs. LUSC). All GNN-based models exceeded an accuracy of 70\%, with GIN reaching 74.5\% (Table \ref{table1}), close to the ProteinPNet ablation with random prototypes. Next, we evaluated different post-hoc graph explainers, namely GNNExplainer \cite{ying2019gnnexplainer}, GNN-LRP \cite{schwarzenberg2019layerwise} and GraphGrad-CAM \cite{pope2019explainability} (Figure \ref{learned_prototypes}B using the trained GIN model. We observed that GNNExplainer and GNN-LRP highlighted cells and nodes that are randomly distributed in the tissue (Figure \ref{learned_prototypes}). Conversely, GraphGrad-CAM, resulted in spatially colocalized high/low importance regions, occasionally antithetical than those of ProteinPNet. We then estimated how well these graph explainers agree with each other and with a random explainer (Figure \ref{jaccard_matrix}) and observed consistently low average scores, suggesting that pairwise agreement between methods is on par with that of a random selection of cells. This result further highlights the limitations of post-hoc explainers: although GNN baselines can predict the disease subtype, the underlying high-importance cells are likely capturing spatially meaningful patterns. 
%

\textbf{Prototype interpretation}
We conducted an exploratory analysis as part of the workflow to identify biological concepts encapsulated in the prototypes. For each prototype, we selected the $m=100$ most prototypically activated samples and isolated regions above the 80th percentile of activation as prototypical regions, with the remaining regions in the same samples serving as references, and investigated differences among prototypes and between prototypes and references.
Qualitatively, the 100 LUAD and LUSC highest activation images differ in both tissue morphology and spatial heterogeneity (Figure \ref{lusc_prototypes} and Figure \ref{luad_prototypes}. LUSC prototypes appear as clumps of densely connected, small regions with minimal infiltration, whereas LUAD prototypes exhibit pronounced glandular morphology. These prototypes align with the typical LUAD and LUSC core patterns (see Fig. 2A of \cite{cords_cancer-associated_2024}). 
To quantify those differences, we estimated different heterogeneity metrics in prototypically activated regions using ATHENA \cite{10.1093/bioinformatics/btac303}. To test differences in cell density and topology, we compared extent (the ratio of the tissue area to the area of its bounding box) and coreness (the maximal $k$-node subgraph with nodes of degree $\geq k$) between prototypes. Both extent ($p < 10^{-3}$) and coreness ($p < 1.03 \cdot 10^{-4}$) were higher in LUSC prototype regions, confirming our hypothesis on connectivity (Figure \ref{graphical_metrics}, Figure \ref{extent}). To assess tumor infiltration, we used the infiltration score in ATHENA, defined as the ratio of tumor/non-tumor to tumor-tumor interaction. LUAD prototypes showed significantly higher infiltration of non-tumor cells (fibroblasts and immune cells) into tumor regions ( Figure \ref{infilitartion}), consistent with existing literature \cite{cords_cancer-associated_2024}. Finally, when computing modularity \cite{networkx_modularity} based on tumor-nontumor partition, both LUAD and LUSC prototype regions show significantly lower values than reference regions $(p < 10^{-3})$, suggesting that prototypes preferentially capture spatially heterogeneous tumor edges enriched with infiltration and tumor-non-tumor interaction (Figure \ref{modularity}).
\vspace{-5pt}
\begin{center}
\begin{table}[h]
  \caption{Results from all experiments. Averaged over 5 runs.}
  \label{table1}
  \centering
  \begin{tabular}{lll}
    \toprule
    \cmidrule(r){1-2}
    Architecture & Experiment & Accuracy (\%)\\
    \midrule
    ProteinPNet & Synthetic Dataset &  $100\pm 0$\\ 
    GCN    & NSCLC & $71.5 \pm 0.7$      \\
    GIN    & NSCLC & $74.6 \pm 1.4$      \\
    GAT    & NSCLC & $72.3 \pm 1.3$      \\
    ProteinPNet & NSCLC (no cell type information) & $71.2 \pm 0.4 $ \\
    ProteinPNet  & NSCLC (randomized prototypes) & $74.5 \pm 1.5$  \\
    ProteinPNet    & NSCLC & $80.7 \pm 0.4$      \\
    \bottomrule
  \end{tabular}
\end{table}
\end{center}

\vspace{-30pt}
\section{Conclusions}
\vspace{-10pt}
In this work, we present ProteinPNet, a prototypical part-network-based framework for learning and analyzing spatial motifs from spatial proteomics data. 
On synthetic data, ProteinPNet reliably recovered ground-truth class-specific prototypes, while on the NSCLC dataset, it achieved high predictive accuracy and identified prototypes consistent with known LUAD and LUSC patterns. Ablations confirmed that both the prototype learning process and protein expression contribute critically to performance. Downstream analyses further verified that prototypes capture biologically meaningful patterns of tissue connectivity and tumor infiltration. Together, these results demonstrate the potential of prototype-based learning to uncover recurrent organizational structures in TME and to provide interpretable insights into cancer biology.
A main limitation of the current implementation of our model is the use of PCA to compress the 43-plex images to a pseudo-RGB representation. Although training the model on the PCA-reduced images led to high prediction accuracy and was able to capture high-level spatial motifs, it potentially caused some loss of information from the original data. We are currently investigating architectural modifications and more powerful encoding strategies that can better preserve the richness and complexity of multiplexed images. This is particularly important in view of future extension of ProteinPNet to spatial transcriptomics data that contain thousands of channels. In addition, we are exploring best practices for analyzing and interpreting spatial motifs to yield more refined biological insights, as the optimal way to interpret and use these prototypical spatial motifs is currently guided by visual inspections.

\textbf{Disclosure of Funding}
This project has been made possible in part by grant numbers 202297, 215550 and 235972 from the Swiss National Science Foundation and grant number 2024-345909 from the Chan Zuckerberg Initiative DAF, an advised fund of Silicon Valley Community Foundation.

\newpage
\bibliographystyle{unsrtnat}
\bibliography{neurips_2025}

\medskip


\appendix
\renewcommand{\thefigure}{S\arabic{figure}}
\setcounter{figure}{0}

\section{Supplementary Figures}

\subsection{Synthetic Data Supplementary Figures}

\begin{figure}[H]
\includegraphics[width=\textwidth]{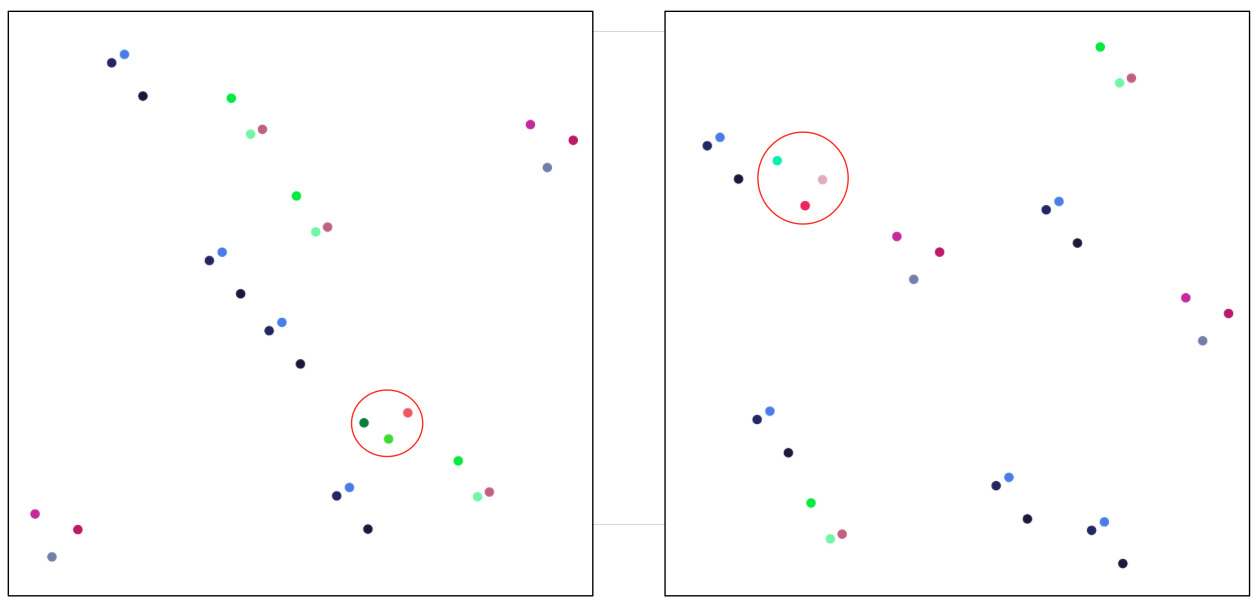} 
\caption{The two classes present in the synthetic dataset, with the red circle outlining the class-defining prototypes. Class independent prototypes can be seen in both samples.} 
\label{supp:synth_data}
\end{figure}

\begin{figure}[H]
\includegraphics[width=\textwidth]{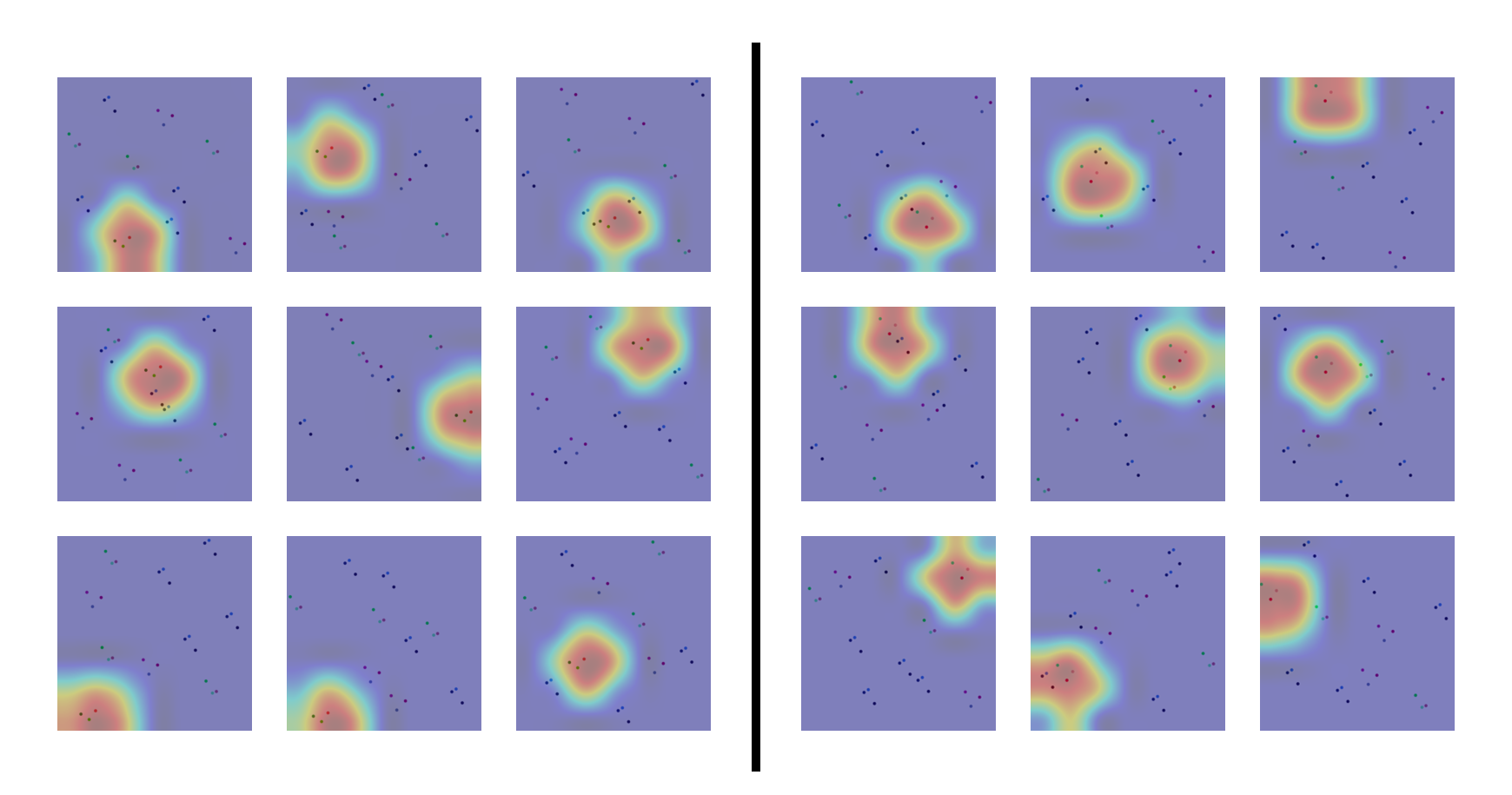}
\caption{Synthetic data activations.} \label{synth_data_activations}
\end{figure}

\begin{figure}[H]
\includegraphics[width=\textwidth]{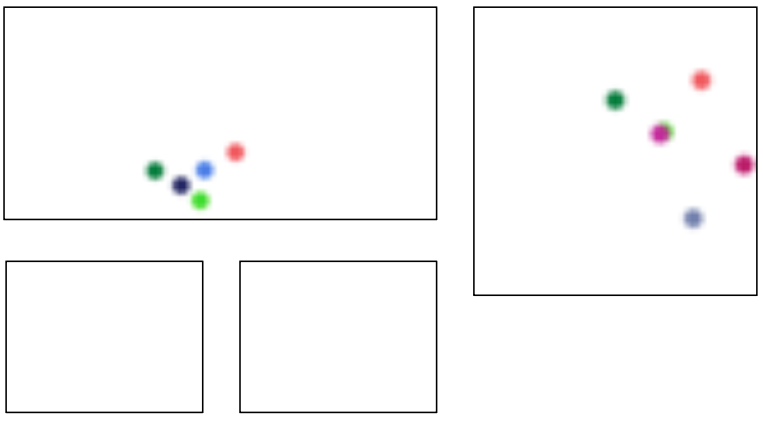}
\caption{Example of prototypes extracted from the synthetic dataset as above. These are class specific prototypes, indicating that each of the top two prototypes belong to the first class and the bottom two belong to the second class. One can see that the second prototype contains an occlusion of a neutral prototype over the classifying prototype. In order to demonstrate the lack of the classifying prototype in the other class, the model focuses on the white space present in the model.} \label{extracted_prototypes}
\end{figure}

\subsection{Top-100 Prototypically Activated Images Supplementary Figures}

\begin{figure}[H]
\includegraphics[width=\textwidth]{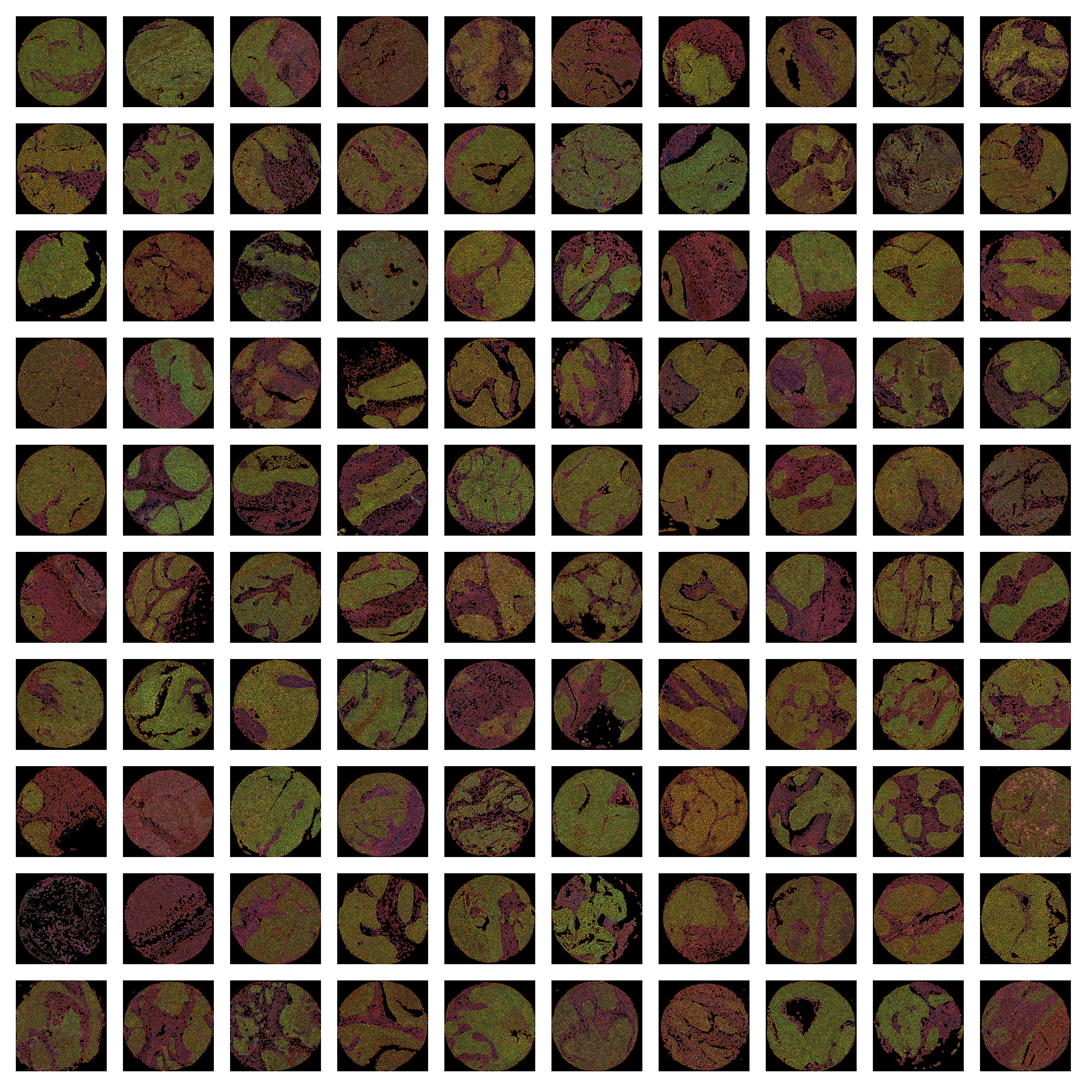}
\caption{100 images with highest activation to LUSC prototype in model with highest accuracy.} \label{lusc_prototypes}
\end{figure}

\begin{figure}[H]
\includegraphics[width=\textwidth]{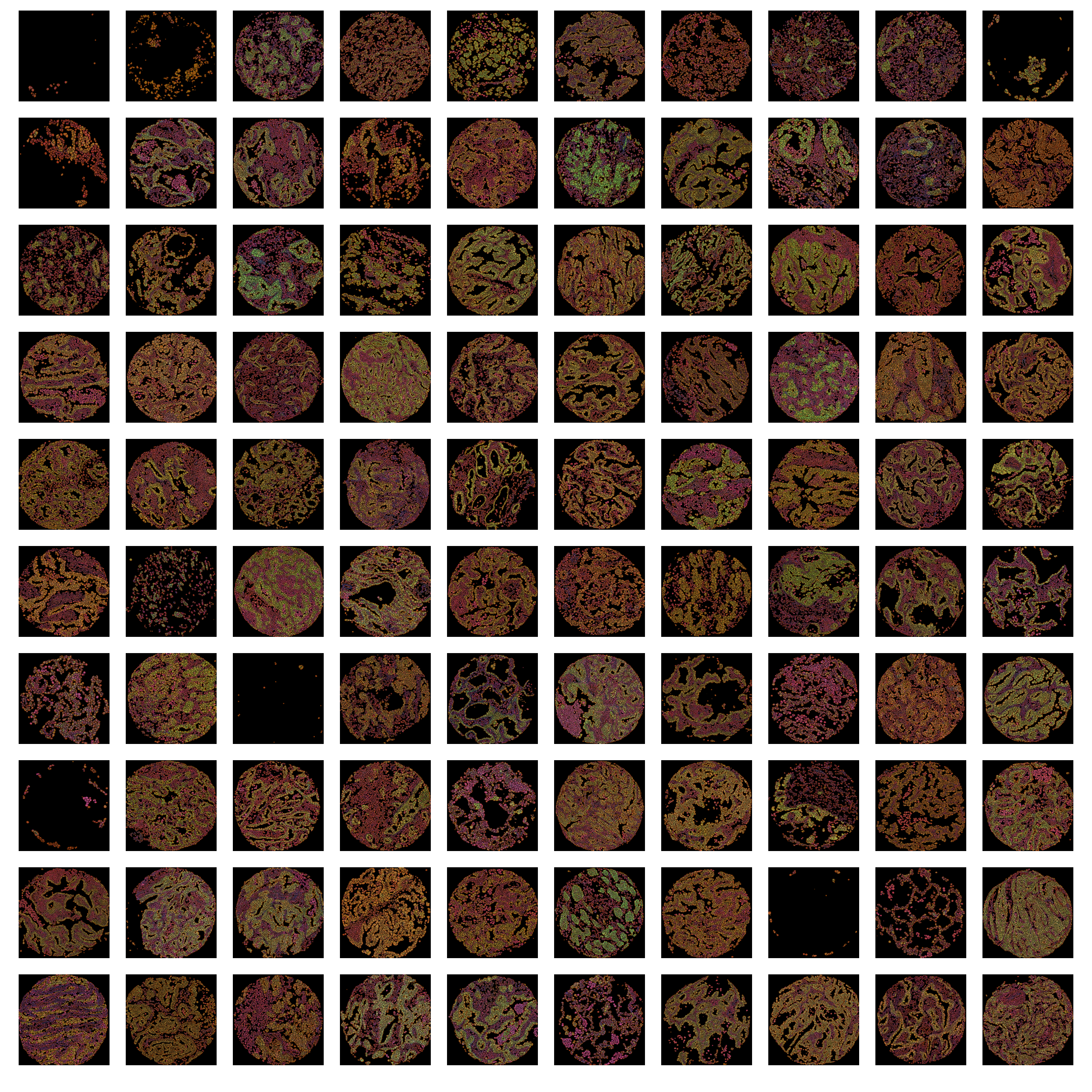}
\caption{100 images with highest activation to LUAD prototype in model with highest accuracy.} \label{luad_prototypes}
\end{figure}

\begin{figure}[H]
\centering
\includegraphics[width=0.75\textwidth]{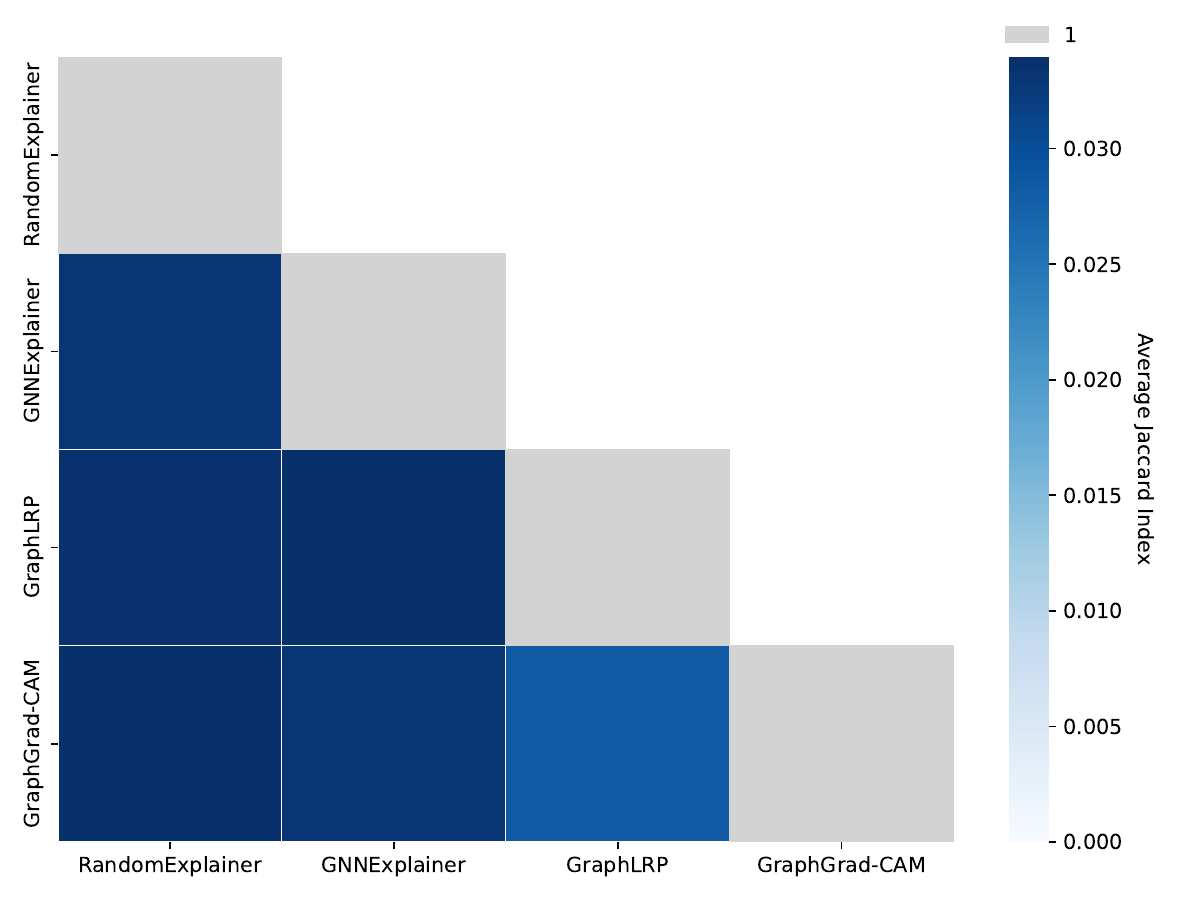} 
\caption{Jaccard index between RandomExplainer, GNNExplainer, GraphLRP, GraphGrad-CAM explainers computed for the top 100 most relevant cells. Jaccard index is defined as follows: $J(A_i, A_j) = \frac{|A_i \cap A_j|}{|A_i \cup A_j|}$, where $A_i$ and $A_j$ are set of the top 100 most relevant cells for explainer $i$ and $j$.} \label{jaccard_matrix}
\end{figure}

\subsection{Graphical and Morphological Metrics: Full Results}

\begin{figure}[H]
\includegraphics[width=\textwidth]{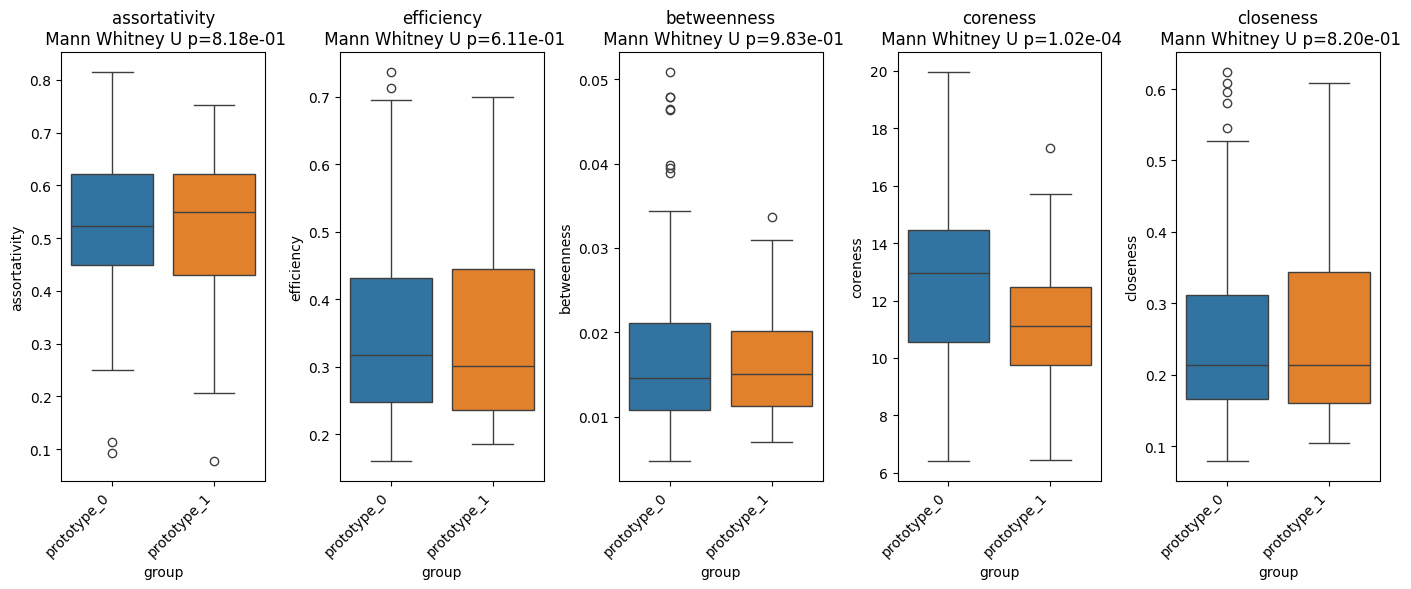}
\caption{Graphical metrics from ATHENA. We can see a statistically significant difference in coreness higher in prototype 0, corresponding to LUSC.} \label{graphical_metrics}
\end{figure}

\begin{figure}[H]
\includegraphics[width=\textwidth]{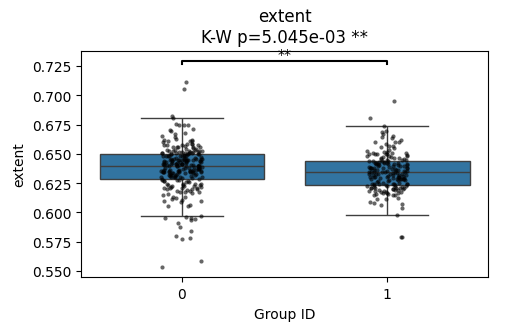}
\caption{Extent by prototype. We can see a statistically significant difference in extent higher in prototype 0, corresponding to LUSC, corresponding to a lower amount of empty space morphologically.} \label{extent}
\end{figure}

\begin{figure}[H]
\includegraphics[width=\textwidth]{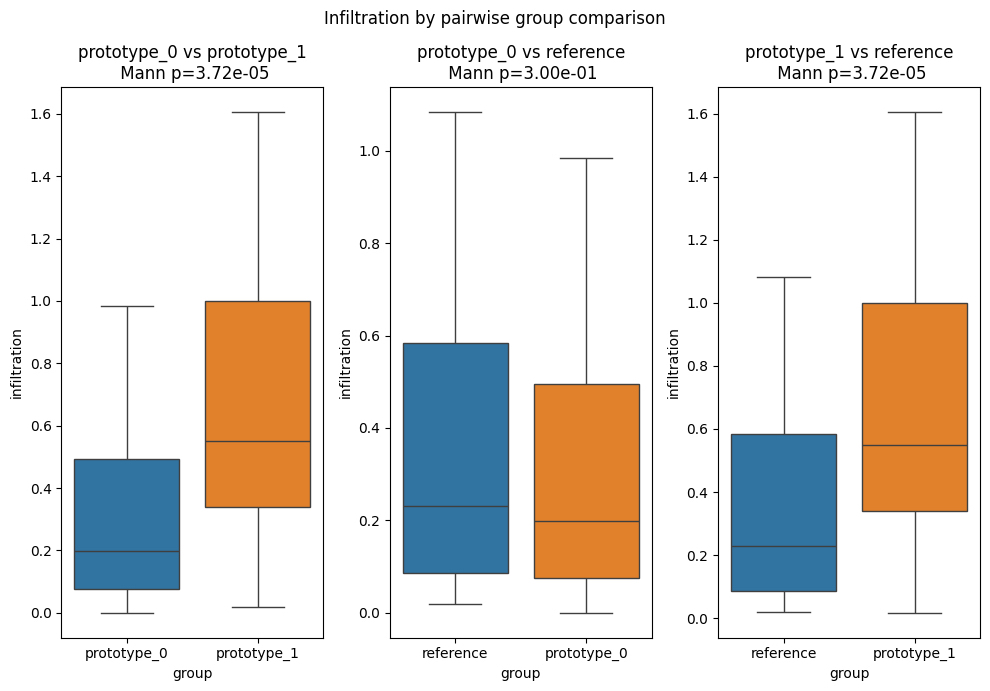}
\caption{Infiltration by prototype. Prototype 0 corresponds to LUSC and prototype 1 corresponds to LUAD. We can see a clear increase in tumor infiltration in prototypes corresponding to LUAD as observed qualitatively.} \label{infilitartion}
\end{figure}

\begin{figure}[H]
\includegraphics[width=\textwidth]{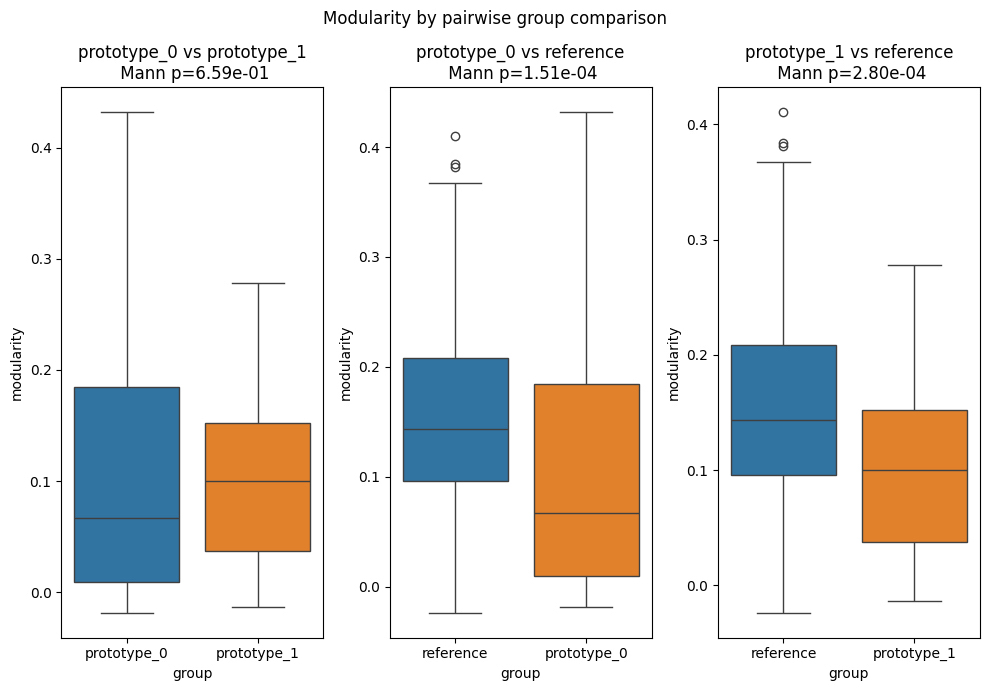}
\caption{Modularity by prototype. Prototype 0 corresponds to LUSC and prototype 1 corresponds to LUAD. We can see that both protototypes express a lower modularity than the reference population, suggesting that heterogeneous regions in the tumor microenvironment are particularly relevant for NSCLC classification.} \label{modularity}
\end{figure}


\newpage

\section{Training Details}

All experiments were conducted using a ResNet152 backbone with a 60/20/20 train/test/val split on PCA reduced data. Hyperparameters were borrowed directly from the ProtoPNet paper. In all experiments 1 prototype was used per class (2 total). All models are trained with ADAM and a StepLR scheduler. More details about training setups can be seen in the code.

\section{Dataset Preprocessing Details}

We have followed the preprocessing of \cite{cords_cancer-associated_2024} for the NSCLC data. We first filter for only samples corresponding to LUAD or LUSC. Of the 43 proteins measured, we first remove both Iridium channels, leaving 41 remaining proteins.

For normalization, the samples are arcsinh transformed before being clipped at the 99th percentile and 0-1 transformed using the global minimum and maximum per channel. Following this, the PCA per pixel is taken over all samples in the train set. This PCA reduction is then applied to the entire dataset, and the dataset is then 0-1 normalized again.

\begin{figure}[H]
\includegraphics[width=\textwidth]{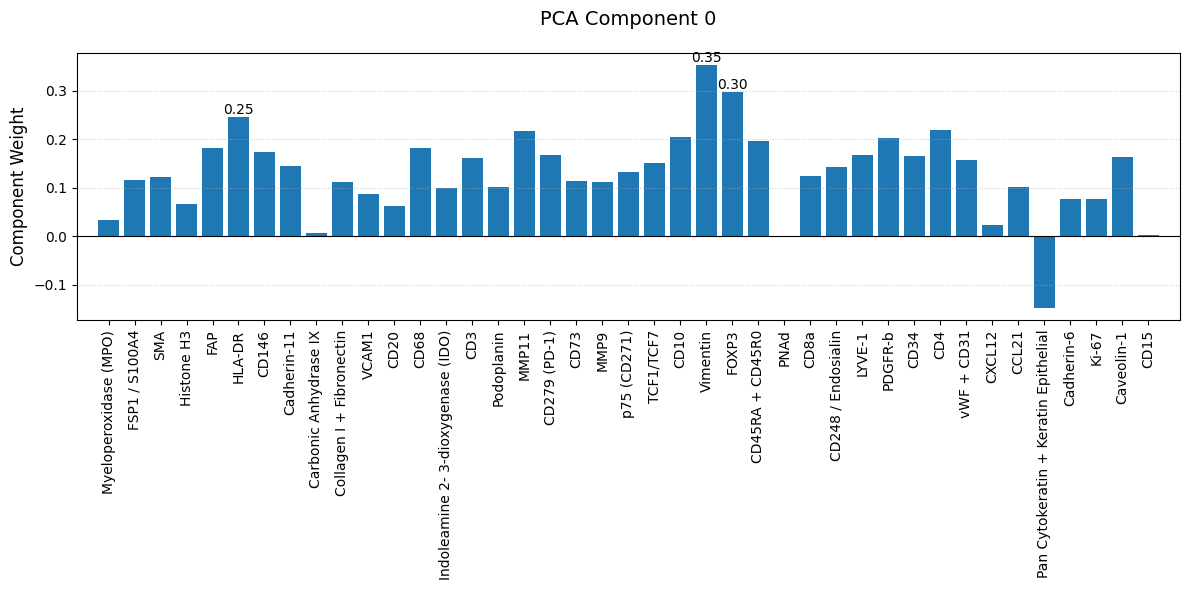}\label{pca}
\caption{PCA Component 0 (Stromal Cells)} \label{pca0}
\end{figure}
\begin{figure}[H]
\includegraphics[width=\textwidth]{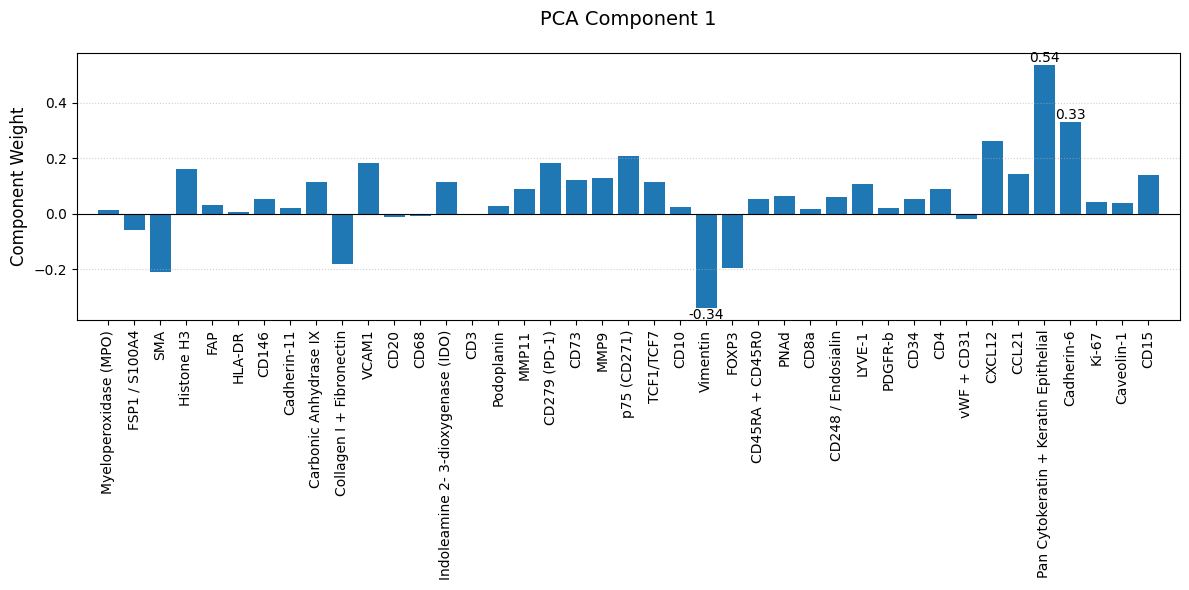}
\caption{PCA Component 1 (Epithelial Cells)} \label{pca1}
\end{figure}
\begin{figure}[H]
\includegraphics[width=\textwidth]{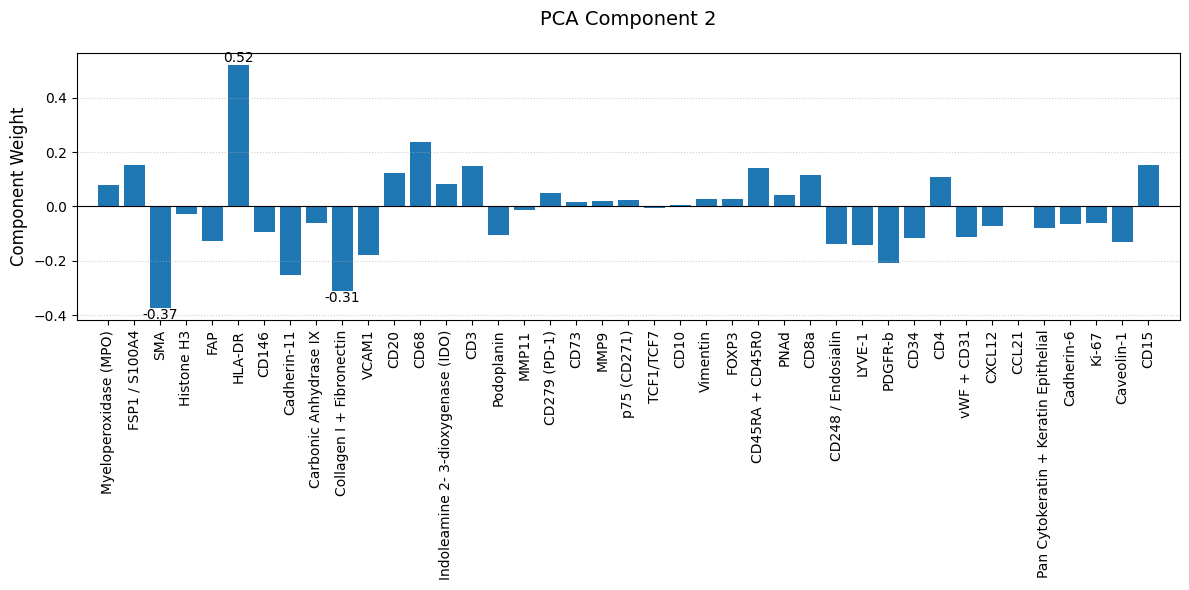}
\caption{PCA Component 2 (Immune Cells)} \label{pca2}
\end{figure}
\begin{figure}[H]
\includegraphics[width=\textwidth]{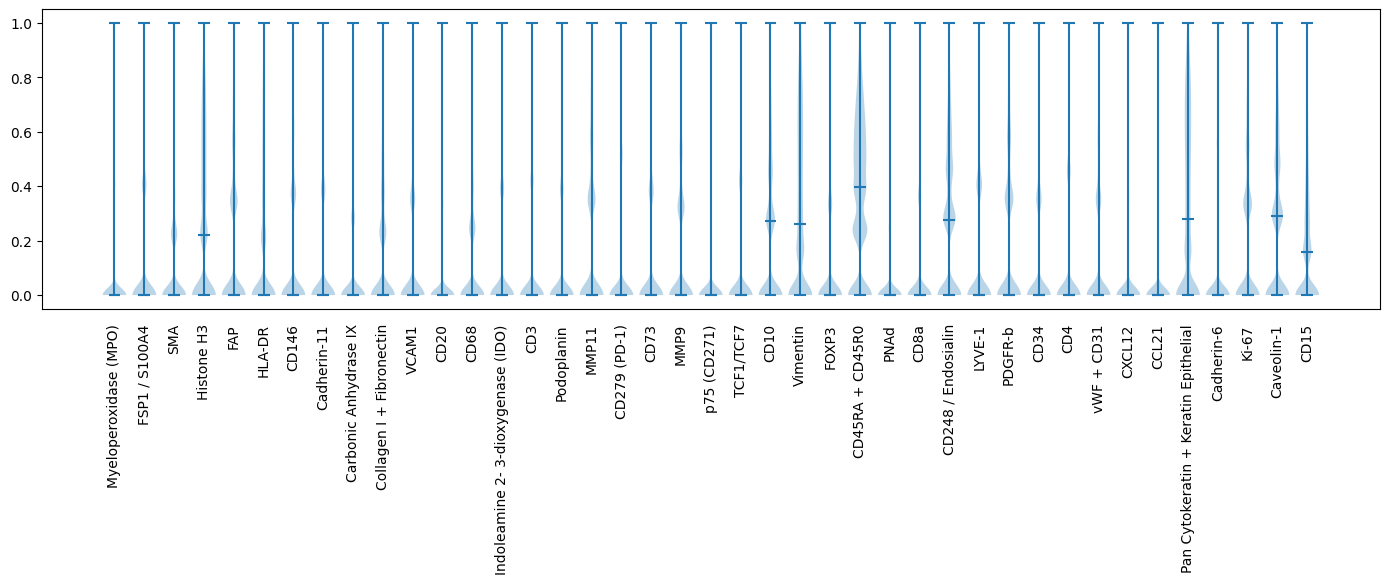}
\caption{Violin plot of protein expression distributions over all cells in the NSCLC data.} \label{violin}
\end{figure}

\section*{NeurIPS Paper Checklist}


\begin{enumerate}

\item {\bf Claims}
    \item[] Question: Do the main claims made in the abstract and introduction accurately reflect the paper's contributions and scope?
    \item[] Answer: \answerYes{}{} 
    \item[] Justification: We discuss the limitations of post-hoc interpretability in spatial omics and identify a literature gap in interpretable methods for concept discovery in spatial omics, all of which are addressed in the main paper.
    \item[] Guidelines:
    \begin{itemize}
        \item The answer NA means that the abstract and introduction do not include the claims made in the paper.
        \item The abstract and/or introduction should clearly state the claims made, including the contributions made in the paper and important assumptions and limitations. A No or NA answer to this question will not be perceived well by the reviewers. 
        \item The claims made should match theoretical and experimental results, and reflect how much the results can be expected to generalize to other settings. 
        \item It is fine to include aspirational goals as motivation as long as it is clear that these goals are not attained by the paper. 
    \end{itemize}

\item {\bf Limitations}
    \item[] Question: Does the paper discuss the limitations of the work performed by the authors?
    \item[] Answer: \answerYes{} 
    \item[] Justification: We contain a limitations section in the conclusion addressing future work and limitations of this existing, preliminary approach to prototype discovery in spatial omics. 
    \item[] Guidelines:
    \begin{itemize}
        \item The answer NA means that the paper has no limitation while the answer No means that the paper has limitations, but those are not discussed in the paper. 
        \item The authors are encouraged to create a separate "Limitations" section in their paper.
        \item The paper should point out any strong assumptions and how robust the results are to violations of these assumptions (e.g., independence assumptions, noiseless settings, model well-specification, asymptotic approximations only holding locally). The authors should reflect on how these assumptions might be violated in practice and what the implications would be.
        \item The authors should reflect on the scope of the claims made, e.g., if the approach was only tested on a few datasets or with a few runs. In general, empirical results often depend on implicit assumptions, which should be articulated.
        \item The authors should reflect on the factors that influence the performance of the approach. For example, a facial recognition algorithm may perform poorly when image resolution is low or images are taken in low lighting. Or a speech-to-text system might not be used reliably to provide closed captions for online lectures because it fails to handle technical jargon.
        \item The authors should discuss the computational efficiency of the proposed algorithms and how they scale with dataset size.
        \item If applicable, the authors should discuss possible limitations of their approach to address problems of privacy and fairness.
        \item While the authors might fear that complete honesty about limitations might be used by reviewers as grounds for rejection, a worse outcome might be that reviewers discover limitations that aren't acknowledged in the paper. The authors should use their best judgment and recognize that individual actions in favor of transparency play an important role in developing norms that preserve the integrity of the community. Reviewers will be specifically instructed to not penalize honesty concerning limitations.
    \end{itemize}

\item {\bf Theory assumptions and proofs}
    \item[] Question: For each theoretical result, does the paper provide the full set of assumptions and a complete (and correct) proof?
    \item[] Answer: \answerNA{} 
    \item[] Justification: there are no theoretical results. 
    \item[] Guidelines:
    \begin{itemize}
        \item The answer NA means that the paper does not include theoretical results. 
        \item All the theorems, formulas, and proofs in the paper should be numbered and cross-referenced.
        \item All assumptions should be clearly stated or referenced in the statement of any theorems.
        \item The proofs can either appear in the main paper or the supplemental material, but if they appear in the supplemental material, the authors are encouraged to provide a short proof sketch to provide intuition. 
        \item Inversely, any informal proof provided in the core of the paper should be complemented by formal proofs provided in appendix or supplemental material.
        \item Theorems and Lemmas that the proof relies upon should be properly referenced. 
    \end{itemize}

    \item {\bf Experimental result reproducibility}
    \item[] Question: Does the paper fully disclose all the information needed to reproduce the main experimental results of the paper to the extent that it affects the main claims and/or conclusions of the paper (regardless of whether the code and data are provided or not)?
    \item[] Answer \answerYes{}
    \item[] Justification: All codes for reproducing the results are published with the paper, and all data can be downloaded online in the link and reference provided in the paper.
    \item[] Guidelines:
    \begin{itemize}
        \item The answer NA means that the paper does not include experiments.
        \item If the paper includes experiments, a No answer to this question will not be perceived well by the reviewers: Making the paper reproducible is important, regardless of whether the code and data are provided or not.
        \item If the contribution is a dataset and/or model, the authors should describe the steps taken to make their results reproducible or verifiable. 
        \item Depending on the contribution, reproducibility can be accomplished in various ways. For example, if the contribution is a novel architecture, describing the architecture fully might suffice, or if the contribution is a specific model and empirical evaluation, it may be necessary to either make it possible for others to replicate the model with the same dataset, or provide access to the model. In general. releasing code and data is often one good way to accomplish this, but reproducibility can also be provided via detailed instructions for how to replicate the results, access to a hosted model (e.g., in the case of a large language model), releasing of a model checkpoint, or other means that are appropriate to the research performed.
        \item While NeurIPS does not require releasing code, the conference does require all submissions to provide some reasonable avenue for reproducibility, which may depend on the nature of the contribution. For example
        \begin{enumerate}
            \item If the contribution is primarily a new algorithm, the paper should make it clear how to reproduce that algorithm.
            \item If the contribution is primarily a new model architecture, the paper should describe the architecture clearly and fully.
            \item If the contribution is a new model (e.g., a large language model), then there should either be a way to access this model for reproducing the results or a way to reproduce the model (e.g., with an open-source dataset or instructions for how to construct the dataset).
            \item We recognize that reproducibility may be tricky in some cases, in which case authors are welcome to describe the particular way they provide for reproducibility. In the case of closed-source models, it may be that access to the model is limited in some way (e.g., to registered users), but it should be possible for other researchers to have some path to reproducing or verifying the results.
        \end{enumerate}
    \end{itemize}

\item {\bf Open access to data and code}
    \item[] Question: Does the paper provide open access to the data and code, with sufficient instructions to faithfully reproduce the main experimental results, as described in supplemental material?
    \item[] Answer: \answerYes{}{} 
    \item[] Justification: We open source all code and data
    \item[] Guidelines:
    \begin{itemize}
        \item The answer NA means that paper does not include experiments requiring code.
        \item Please see the NeurIPS code and data submission guidelines (\url{https://nips.cc/public/guides/CodeSubmissionPolicy}) for more details.
        \item While we encourage the release of code and data, we understand that this might not be possible, so “No” is an acceptable answer. Papers cannot be rejected simply for not including code, unless this is central to the contribution (e.g., for a new open-source benchmark).
        \item The instructions should contain the exact command and environment needed to run to reproduce the results. See the NeurIPS code and data submission guidelines (\url{https://nips.cc/public/guides/CodeSubmissionPolicy}) for more details.
        \item The authors should provide instructions on data access and preparation, including how to access the raw data, preprocessed data, intermediate data, and generated data, etc.
        \item The authors should provide scripts to reproduce all experimental results for the new proposed method and baselines. If only a subset of experiments are reproducible, they should state which ones are omitted from the script and why.
        \item At submission time, to preserve anonymity, the authors should release anonymized versions (if applicable).
        \item Providing as much information as possible in supplemental material (appended to the paper) is recommended, but including URLs to data and code is permitted.
    \end{itemize}

\item {\bf Experimental setting/details}
    \item[] Question: Does the paper specify all the training and test details (e.g., data splits, hyperparameters, how they were chosen, type of optimizer, etc.) necessary to understand the results?
    \item[] Answer: \answerYes{} 
    \item[] Justification: All preprocessing information and reproducibility information can be found in the methods section, appendix, or in the code.
    \item[] Guidelines:
    \begin{itemize}
        \item The answer NA means that the paper does not include experiments.
        \item The experimental setting should be presented in the core of the paper to a level of detail that is necessary to appreciate the results and make sense of them.
        \item The full details can be provided either with the code, in appendix, or as supplemental material.
    \end{itemize}

\item {\bf Experiment statistical significance}
    \item[] Question: Does the paper report error bars suitably and correctly defined or other appropriate information about the statistical significance of the experiments?
    \item[] Answer: \answerYes{} 
    \item[] Justification: All accuracy metrics have standard error, and box plots and p values are provided for all graphical and morphological analyses.
    \item[] Guidelines:
    \begin{itemize}
        \item The answer NA means that the paper does not include experiments.
        \item The authors should answer "Yes" if the results are accompanied by error bars, confidence intervals, or statistical significance tests, at least for the experiments that support the main claims of the paper.
        \item The factors of variability that the error bars are capturing should be clearly stated (for example, train/test split, initialization, random drawing of some parameter, or overall run with given experimental conditions).
        \item The method for calculating the error bars should be explained (closed form formula, call to a library function, bootstrap, etc.)
        \item The assumptions made should be given (e.g., Normally distributed errors).
        \item It should be clear whether the error bar is the standard deviation or the standard error of the mean.
        \item It is OK to report 1-sigma error bars, but one should state it. The authors should preferably report a 2-sigma error bar than state that they have a 96\% CI, if the hypothesis of Normality of errors is not verified.
        \item For asymmetric distributions, the authors should be careful not to show in tables or figures symmetric error bars that would yield results that are out of range (e.g. negative error rates).
        \item If error bars are reported in tables or plots, The authors should explain in the text how they were calculated and reference the corresponding figures or tables in the text.
    \end{itemize}

\item {\bf Experiments compute resources}
    \item[] Question: For each experiment, does the paper provide sufficient information on the computer resources (type of compute workers, memory, time of execution) needed to reproduce the experiments?
    \item[] Answer: \answerYes{}
    \item[] Justification: The work uses publicly available datasets \cite{cords_cancer-associated_2024} and synthetic data, with no personally identifiable information. All experiments comply with standard practices in computational biology and machine learning, and there are no ethical concerns related to data collection, usage, or reporting. The research was conducted in accordance with the NeurIPS Code of Ethics.
    \item[] Guidelines:
    \begin{itemize}
        \item The answer NA means that the paper does not include experiments.
        \item The paper should indicate the type of compute workers CPU or GPU, internal cluster, or cloud provider, including relevant memory and storage.
        \item The paper should provide the amount of compute required for each of the individual experimental runs as well as estimate the total compute. 
        \item The paper should disclose whether the full research project required more compute than the experiments reported in the paper (e.g., preliminary or failed experiments that didn't make it into the paper). 
    \end{itemize}
    
\item {\bf Code of ethics}
    \item[] Question: Does the research conducted in the paper conform, in every respect, with the NeurIPS Code of Ethics \url{https://neurips.cc/public/EthicsGuidelines}?
    \item[] Answer: \answerYes{} 
    \item[] Justification: We conform to the code of ethics. 
    \item[] Guidelines:
    \begin{itemize}
        \item The answer NA means that the authors have not reviewed the NeurIPS Code of Ethics.
        \item If the authors answer No, they should explain the special circumstances that require a deviation from the Code of Ethics.
        \item The authors should make sure to preserve anonymity (e.g., if there is a special consideration due to laws or regulations in their jurisdiction).
    \end{itemize}

\item {\bf Broader impacts}
    \item[] Question: Does the paper discuss both potential positive societal impacts and negative societal impacts of the work performed?
    \item[] Answer: \answerNo{}
    \item[] Justification: Our paper focuses on methodological contributions and empirical validation. While in the abstract and conclusion we highlight potential benefits for precision oncology and interpretable spatial biomarkers, there is no explicit consideration of risks (e.g., misuse, bias, privacy) or mitigation strategies.
    \item[] Guidelines:
    \begin{itemize}
        \item The answer NA means that there is no societal impact of the work performed.
        \item If the authors answer NA or No, they should explain why their work has no societal impact or why the paper does not address societal impact.
        \item Examples of negative societal impacts include potential malicious or unintended uses (e.g., disinformation, generating fake profiles, surveillance), fairness considerations (e.g., deployment of technologies that could make decisions that unfairly impact specific groups), privacy considerations, and security considerations.
        \item The conference expects that many papers will be foundational research and not tied to particular applications, let alone deployments. However, if there is a direct path to any negative applications, the authors should point it out. For example, it is legitimate to point out that an improvement in the quality of generative models could be used to generate deepfakes for disinformation. On the other hand, it is not needed to point out that a generic algorithm for optimizing neural networks could enable people to train models that generate Deepfakes faster.
        \item The authors should consider possible harms that could arise when the technology is being used as intended and functioning correctly, harms that could arise when the technology is being used as intended but gives incorrect results, and harms following from (intentional or unintentional) misuse of the technology.
        \item If there are negative societal impacts, the authors could also discuss possible mitigation strategies (e.g., gated release of models, providing defenses in addition to attacks, mechanisms for monitoring misuse, mechanisms to monitor how a system learns from feedback over time, improving the efficiency and accessibility of ML).
    \end{itemize}
    
\item {\bf Safeguards}
    \item[] Question: Does the paper describe safeguards that have been put in place for responsible release of data or models that have a high risk for misuse (e.g., pretrained language models, image generators, or scraped datasets)?
    \item[] Answer: \answerNA{} 
    \item[] Justification: There are no suck risks for responsible release of data.
    \item[] Guidelines:
    \begin{itemize}
        \item The answer NA means that the paper poses no such risks.
        \item Released models that have a high risk for misuse or dual-use should be released with necessary safeguards to allow for controlled use of the model, for example by requiring that users adhere to usage guidelines or restrictions to access the model or implementing safety filters. 
        \item Datasets that have been scraped from the Internet could pose safety risks. The authors should describe how they avoided releasing unsafe images.
        \item We recognize that providing effective safeguards is challenging, and many papers do not require this, but we encourage authors to take this into account and make a best faith effort.
    \end{itemize}

\item {\bf Licenses for existing assets}
    \item[] Question: Are the creators or original owners of assets (e.g., code, data, models), used in the paper, properly credited and are the license and terms of use explicitly mentioned and properly respected?
    \item[] Answer: \answerYes{} 
    \item[] Justification: The only code used is the ProtoPNet code, and credit is explicitly given (MIT License). We also use the CORDS dataset for the lung cancer data, which is publicly available deposited in \href{https://zenodo.org/records/7961844}{zenodo} under a Creative Commons Attribution 4.0 International license.
    \item[] Guidelines:
    \begin{itemize}
        \item The answer NA means that the paper does not use existing assets.
        \item The authors should cite the original paper that produced the code package or dataset.
        \item The authors should state which version of the asset is used and, if possible, include a URL.
        \item The name of the license (e.g., CC-BY 4.0) should be included for each asset.
        \item For scraped data from a particular source (e.g., website), the copyright and terms of service of that source should be provided.
        \item If assets are released, the license, copyright information, and terms of use in the package should be provided. For popular datasets, \url{paperswithcode.com/datasets} has curated licenses for some datasets. Their licensing guide can help determine the license of a dataset.
        \item For existing datasets that are re-packaged, both the original license and the license of the derived asset (if it has changed) should be provided.
        \item If this information is not available online, the authors are encouraged to reach out to the asset's creators.
    \end{itemize}

\item {\bf New assets}
    \item[] Question: Are new assets introduced in the paper well documented and is the documentation provided alongside the assets?
    \item[] Answer: \answerYes{}{} 
    \item[] Justification: Documentation for synthetic data, preprocessing, etc. is all available in the supplementary zip file in the code.
    \item[] Guidelines:
    \begin{itemize}
        \item The answer NA means that the paper does not release new assets.
        \item Researchers should communicate the details of the dataset/code/model as part of their submissions via structured templates. This includes details about training, license, limitations, etc. 
        \item The paper should discuss whether and how consent was obtained from people whose asset is used.
        \item At submission time, remember to anonymize your assets (if applicable). You can either create an anonymized URL or include an anonymized zip file.
    \end{itemize}

\item {\bf Crowdsourcing and research with human subjects}
    \item[] Question: For crowdsourcing experiments and research with human subjects, does the paper include the full text of instructions given to participants and screenshots, if applicable, as well as details about compensation (if any)? 
    \item[] Answer: \answerNA{}.
    \item[] Justification: The paper does not involve crowdsourcing nor research with human subjects.
    \item[] Guidelines:
    \begin{itemize}
        \item The answer NA means that the paper does not involve crowdsourcing nor research with human subjects.
        \item Including this information in the supplemental material is fine, but if the main contribution of the paper involves human subjects, then as much detail as possible should be included in the main paper. 
        \item According to the NeurIPS Code of Ethics, workers involved in data collection, curation, or other labor should be paid at least the minimum wage in the country of the data collector. 
    \end{itemize}

\item {\bf Institutional review board (IRB) approvals or equivalent for research with human subjects}
    \item[] Question: Does the paper describe potential risks incurred by study participants, whether such risks were disclosed to the subjects, and whether Institutional Review Board (IRB) approvals (or an equivalent approval/review based on the requirements of your country or institution) were obtained?
    \item[] Answer: \answerNA 
    \item[] Justification: no data was collected in this paper. 
    \item[] Guidelines:
    \begin{itemize}
        \item The answer NA means that the paper does not involve crowdsourcing nor research with human subjects.
        \item Depending on the country in which research is conducted, IRB approval (or equivalent) may be required for any human subjects research. If you obtained IRB approval, you should clearly state this in the paper. 
        \item We recognize that the procedures for this may vary significantly between institutions and locations, and we expect authors to adhere to the NeurIPS Code of Ethics and the guidelines for their institution. 
        \item For initial submissions, do not include any information that would break anonymity (if applicable), such as the institution conducting the review.
    \end{itemize}

\item {\bf Declaration of LLM usage}
    \item[] Question: Does the paper describe the usage of LLMs if it is an important, original, or non-standard component of the core methods in this research? Note that if the LLM is used only for writing, editing, or formatting purposes and does not impact the core methodology, scientific rigorousness, or originality of the research, declaration is not required.
    \item[] Answer: \answerNA{} 
    \item[] Justification: No LLMs used.
    \item[] Guidelines:
    \begin{itemize}
        \item The answer NA means that the core method development in this research does not involve LLMs as any important, original, or non-standard components.
        \item Please refer to our LLM policy (\url{https://neurips.cc/Conferences/2025/LLM}) for what should or should not be described.
    \end{itemize}

\end{enumerate}
\end{document}